\documentclass{bmvc2k}

\usepackage{booktabs}
\usepackage{pifont}
\usepackage[table]{xcolor}
\usepackage{multirow}
\usepackage{graphicx}
\usepackage{algorithm}
\usepackage{algorithmic}
\usepackage{amsmath,amssymb,amsfonts}

\title{CoSeP: Complementary Separability Pruning via Class-Separability Clustering}

\addauthor{David Levin}{david.levin2@live.biu.ac.il}{1}
\addauthor{Gonen Singer}{gonen.singer@biu.ac.il}{1}

\addinstitution{
 Faculty of Engineering\\
 Bar-Ilan University\\
 Ramat Gan, Israel
}

\runninghead{Levin, Singer}{Complementary Separability Pruning}

\begin{document}

\maketitle

\begin{abstract}
Neural network pruning aims to compress models for efficient deployment, yet two fundamental challenges remain. First, many methods rely on per-component importance scores, selecting filters or neurons independently and ignoring redundancy: the retained set may include multiple components capturing similar discriminative patterns while missing others entirely. Second, determining per-layer pruning ratios typically requires manual, architecture-specific tuning with no principled stopping criterion.
We propose CoSeP (Complementary Separability Pruning) to address both issues. Rather than scoring components in isolation, CoSeP represents each component by its class-separability profile across all class pairs, computed via Jeffries--Matusita distances. This defines a separability space in which nearby components are potentially redundant and distant components capture complementary information. CoSeP selects a compact set of representatives in this space: components are grouped via $k$-medoids clustering, candidate subset sizes are evaluated using the Mean Simplified Silhouette, and a knee-detection criterion automatically determines how many components to retain. 
Across CIFAR-10, CIFAR-100, and ImageNet-1K, on ResNet, VGG, MobileNet, and DenseNet architectures, CoSeP matches or improves accuracy while reducing FLOPs, with measured wall-clock inference-time reductions of up to $20\%$. For example, it achieves a $+0.66\%$ top-1 accuracy gain with $2.30\times$ FLOPs reduction on ResNet-50/ImageNet-1K, and a $+0.37\%$ gain with $2.59\times$ FLOPs reduction on VGG-16/CIFAR-10. These results demonstrate that modeling complementarity in class-separability space provides an effective and principled approach to pruning.
\end{abstract}

\section{Introduction}
\label{sec:introduction}

Deep neural networks have achieved state-of-the-art performance across a wide range of visual recognition tasks~\cite{liu2022global,he2016deep, simonyan2014very}, yet their high computational demands make deployment on resource-constrained platforms a persistent challenge~\cite{chen2022cicc,han2015deep, he2023structured}. Structured pruning, removing entire components, such as neurons, filters or layers to produce architecturally dense networks, has emerged as a compelling solution, since the resulting models benefit directly from hardware SIMD and tensor-core acceleration without requiring sparse-computation support~\cite{li2016pruning, anwar2017structured}.
Despite substantial progress, two recurring challenges remain.

\textbf{The first concerns which components to prune.} The most widely adopted approaches assign an importance score to each component independently, by weight magnitude, activation statistics, or feature-map rank, and retain the top-scoring components. However, a high-scoring component is not automatically useful given the others retained alongside it. Consider a layer in which the highest-scoring components all respond strongly to the same class boundary: retaining them duplicates the same discriminative information, while components that cover other class pairs are discarded. The fundamental issue is that independent ranking cannot detect inter-component redundancy: two components may each score highly in isolation yet be almost entirely substitutable for one another. Some methods do consider pairwise relationships between components, measuring activation correlation (CHIP~\cite{sui2021chip}) or using control-group statistics (SCOP~\cite{tang2020scop}), but they do not construct an explicit geometry over class-discriminability profiles, and none of the compared methods directly optimizes for a retained set that is collectively complementary: covering diverse discriminative directions across all class pairs simultaneously.

\textbf{The second problem concerns how many components to prune per layer.} Many structured pruning methods require the user to supply a compression target, either globally or per layer, which necessitates expensive trial-and-error sweeps, as the pruning level depends on the interaction between layer characteristics, network architecture, and dataset. In practice, layers differ substantially in their redundancy and sensitivity, making this selection particularly challenging. Methods that automate this step do exist, but typically rely on reinforcement learning~\cite{he2018amc, alwani2022decore}, evolutionary or search-based strategies~\cite{lin2020channel, sun2022aprs}, or auxiliary controller networks~\cite{wu2024auto}, each introducing additional hyperparameters or control variables that must themselves be tuned, effectively shifting rather than eliminating the burden of specifying the pruning level, and requiring substantial additional compute beyond the pruning process itself. A remaining challenge is to derive an intrinsic criterion, grounded in the layer's own redundancy structure, that determines how many components can be removed without relying on a search budget, external objectives, or auxiliary hyperparameter tuning.

We propose CoSeP (Complementary Separability Pruning) to address both problems simultaneously. The central idea is to embed each component into a separability space: each component, a filter in a convolutional layer or a neuron in a linear layer, is represented as a vector of Jeffries--Matusita (JM) distances computed across all class pairs, capturing its ability to discriminate between every pair of class distributions~\cite{kailath1967divergence, dabboor2014jeffries}. For datasets with many classes, this vector is built from all pairwise comparisons among a layer-specific top-$M$ subset of selected classes. Components that occupy nearby positions in this space respond to similar class boundaries and are therefore potentially redundant; those in distant regions cover complementary boundaries and should be jointly retained. Applying \emph{k}-medoids clustering over this space groups similar components together~\cite{schubert2019faster}; selecting the highest-weight representative from each cluster yields a compact, diversity-aware retained set. Crucially, the Mean Simplified Silhouette (MSS) index computed over candidate cluster counts~\cite{levin2024gb, levin2025graph} traces a curve with a natural knee point, detected by the Kneedle algorithm~\cite{satopaa2011finding}, that indicates where additional components cease to meaningfully expand discriminative coverage. This knee directly gives the per-layer retention count: no user-specified compression ratio is needed. The process is applied sequentially across layers, with brief fine-tuning after each step.

\paragraph{Contributions.}
\begin{itemize}
    \item We introduce the \emph{separability embedding space}: a principled representation of each component as its pairwise class-discriminability profile via JM distances. Unlike the compared criteria, namely weight magnitude~\cite{li2016pruning}, activation rank~\cite{lin2020hrank}, pairwise correlation~\cite{sui2021chip}, or filter geometry~\cite{he2019filter}, CoSeP uses a geometry in which distance directly encodes functional redundancy: nearby components separate the same class pairs, while distant components capture complementary discriminative directions. To our knowledge, no prior pruning method constructs or operates in such a geometry.
    \item We propose CoSeP, which combines $k$-medoids clustering over this space with MSS-guided knee detection to automatically determine the per-layer retention count, no manual compression ratio, search budget, or auxiliary network is required. The per-layer pruning ratio emerges directly from the intrinsic redundancy structure of each layer's separability space, rather than from an external search objective.
    \item We evaluate CIFAR-10, CIFAR-100, and ImageNet-1K on six architectures: ResNet-56, VGG-16/19, MobileNet-V2, DenseNet-40, and ResNet-50. CoSeP consistently matches or improves accuracy while reducing FLOPs and wall-clock inference time.
\end{itemize}

\section{Related Work}
\label{sec:related-work}

\paragraph{Structured pruning and importance criteria.} Structured pruning removes entire components, namely filters in convolutional layers or neurons in linear layers, yielding dense models that benefit directly from hardware acceleration~\cite{anwar2017structured}. The dominant paradigm ranks components by an importance score and removes the lowest-ranked ones. Weight magnitude (L1-norm) is the earliest and still widely-used criterion~\cite{li2016pruning}; Network Slimming~\cite{liu2017learning} achieves a similar effect by regularizing batch-normalization scaling factors. Feature-map rank (HRank~\cite{lin2020hrank}) and filter geometry (FPGM~\cite{he2019filter}) replace weight-based scores with activation statistics; ThiNet~\cite{luo2017thinet} removes filters whose activation contribution to the next layer falls below a learned threshold. More recent methods introduce richer criteria: Torque~\cite{gupta2024torque} applies a physics-inspired regularizer to induce sparsity; SANP~\cite{gao2023sanp} maintains alignment between the pruned and original network via partial regularization; DepGraph~\cite{fang2023depgraph} models inter-layer coupling to prune structurally dependent components jointly. Across all these methods, however, the decision to retain a given component takes no account of the other components retained alongside it, leaving inter-component redundancy unaddressed.

\paragraph{Pairwise and redundancy-aware selection.} A complementary line of work recognizes that components should not be scored in isolation. DCP~\cite{zhuang2018dcp} adds discrimination-aware losses that encourage the pruned network to preserve class-separating directions. CHIP~\cite{sui2021chip} measures pairwise Pearson correlation between components, preferring low-correlation subsets; SCOP~\cite{tang2020scop} compares each component against statistical knockoff features to identify redundant structures; CKA~\cite{pons2024effective} prunes layers whose representations are highly similar to those of the full model. APIB~\cite{guo2023automatic} applies the Information Bottleneck principle via HSIC Lasso to identify globally informative features. SMCP~\cite{humble2022soft} uses soft-mask structured pruning to encourage sparsity while preserving accuracy. CSPrune~\cite{chen2024csprune} enforces class-margin constraints during selection. Unlike CSPrune, CoSeP represents each component by a pairwise class-separability profile and performs clustering in that space, so redundancy reduction is enforced explicitly at the set level rather than indirectly through a discriminative objective. More broadly, although these methods account for inter-component relationships, none of the compared methods constructs an explicit geometry over class-discriminability profiles: they rely on raw activation statistics, similarity measures, or discriminative constraints, rather than modeling how each component separates class pairs. As a result, they cannot directly enforce that the retained set is collectively \emph{complementary}, covering diverse class-discriminative directions simultaneously. CoSeP addresses this gap by embedding components in a class-separability space and applying clustering directly in that geometry.

\paragraph{Automatic compression-ratio determination.} Determining how many components to remove per layer is at least as challenging as choosing which ones to remove. AMC~\cite{he2018amc} frames per-layer ratio search as a reinforcement learning problem; ABCPruner~\cite{lin2020channel} employs an evolutionary algorithm (Artificial Bee Colony) to automatically search for optimal per-layer channel counts; DECORE~\cite{alwani2022decore} uses RL to jointly search structure and compression rates; APRS~\cite{sun2022aprs} searches pruning rates automatically; Autoprune~\cite{xiao2019autoprune} regularizes auxiliary sparsity parameters; AAP~\cite{zhao2023aap} exploits attention maps to infer per-layer ratios; ATO~\cite{wu2024auto} trains a controller network to guide pruning from scratch; GETA~\cite{qu2025geta} extends automatic ratio determination to joint pruning and quantization. Each approach requires either a substantial search budget, an auxiliary network, a custom regularization objective, or additional hyperparameter tuning, ultimately shifting rather than eliminating the cost of specifying the pruning level. CoSeP instead derives each layer's retention count from its MSS curve during one sequential pass through the prunable layers, without reinforcement-learning episodes, evolutionary search, or an auxiliary controller.

\paragraph{Post-hoc versus training-integrated pruning.}
Training-integrated and post-hoc methods address different deployment regimes. REPrune~\cite{park2024reprune} learns channel representatives within a dedicated simultaneous training-and-pruning procedure, whereas CoSeP starts from a trained checkpoint and applies brief layer-wise recovery. CoSeP therefore targets settings in which full training from scratch or reproduction of the original training pipeline is impractical. It still requires labeled calibration data and a small labeled recovery subset, as described in Section~\ref{sec:experimental-setup}; we do not claim data-free pruning.


\begin{algorithm}[t]
\caption{CoSeP: Complementary Separability Pruning}
\label{alg:cosep-pruning-procedure}
\begin{algorithmic}[1]
\REQUIRE Trained network $f(\cdot;\theta)$, calibration set $\mathcal{D}_{\mathrm{cal}}$, prunable layers $\mathcal{L}$
\FOR{each layer $\ell \in \mathcal{L}$}
    \STATE Extract scalar component activations $\{z_{\ell,c}(x)\}$ on $\mathcal{D}_{\mathrm{cal}}$
    \STATE Estimate class-conditional statistics $\mu_{\ell,c}^{(t)}, \sigma_{\ell,c}^{(t)}$
    \STATE Compute JM profiles $\{\mathbf{s}_{\ell,c}\}_{c=1}^{N_\ell}$
    \FOR{candidate $k$}
        \STATE Run $k$-medoids in separability space
        \STATE Compute $\mathrm{MSS}_\ell(k)$
    \ENDFOR
    \STATE Fit a degree-$p$ polynomial to the MSS curve and detect $k_\ell^\star$ with Kneedle
    \STATE Run $k$-medoids with $k=k_\ell^\star$
    \STATE Retain one representative per cluster using Equation~\eqref{eq:cluster-representative}
    \STATE Prune all non-retained components in layer $\ell$
    \STATE Briefly fine-tune the network
\ENDFOR
\RETURN Pruned network
\end{algorithmic}
\end{algorithm}

\begin{figure}[t!]
    \centering
    \includegraphics[width=\textwidth,trim={1.5cm 70cm 2cm 2cm}, clip]{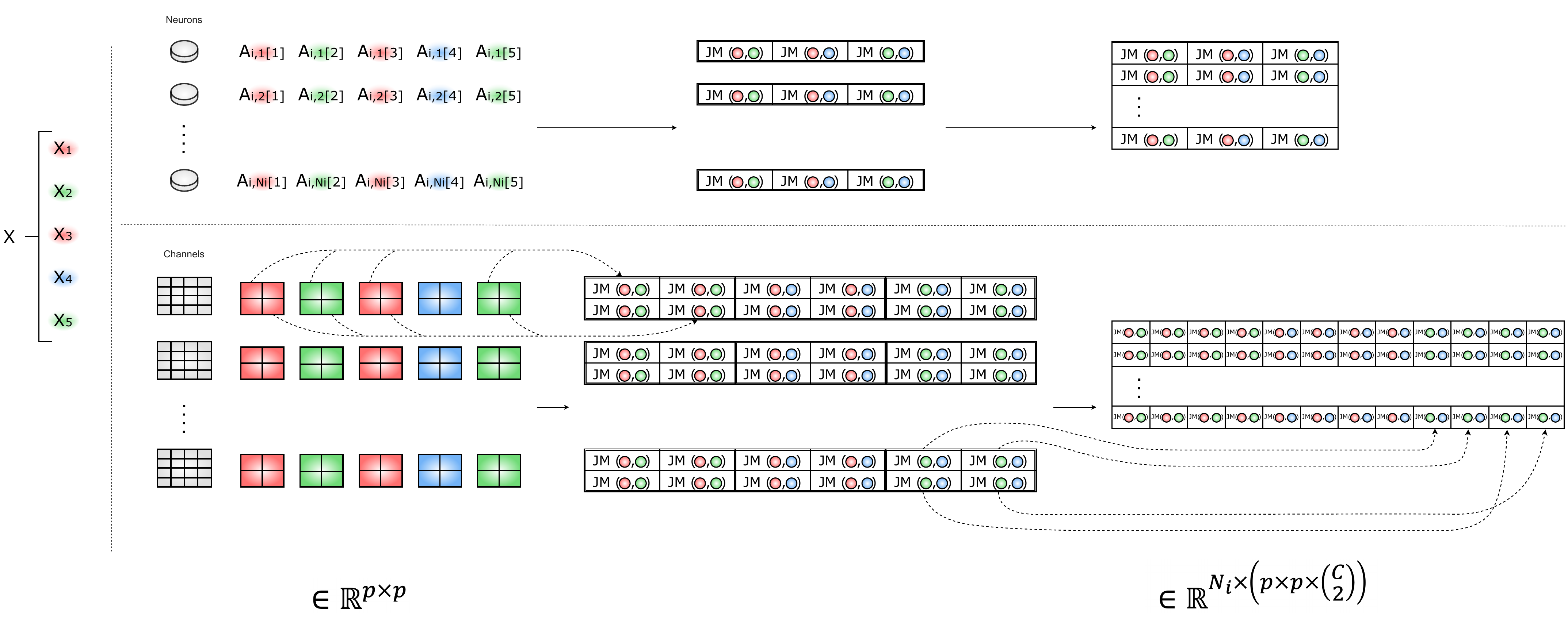}
    \caption{\textbf{Construction of the layer separability matrix used by CoSeP}. For each labeled sample, we extract one scalar activation per structured component: direct activations for linear neurons and spatially averaged activations for convolutional channels. For each component and each selected class pair, we compute a JM separability score, forming a class-separability profile. Stacking these profiles across components yields the layer separability matrix, whose rows are then clustered by CoSeP to select complementary representatives and determine the retained cardinality.}
    \label{fig:separability-matrix-construction}
\end{figure}


\section{Method}
\label{sec:method}

CoSeP is a supervised, post-hoc, layer-wise pruning method for CNN image classifiers. It uses a labeled calibration set to estimate class-conditional activation statistics and a small labeled subset for brief recovery fine-tuning. It does not require training the model from scratch or specifying a target compression ratio, but it is not a data-free pruning method.
For each layer, CoSeP builds a \emph{separability space} in which each component is represented by how well it separates class pairs. Components that lie close in this space tend to encode similar discriminative behavior and are therefore candidates for redundancy, whereas distant components tend to capture complementary discriminative information.
CoSeP then selects one representative from each complementary group and determines the number of groups automatically from the structure of the space itself.
Figure~\ref{fig:separability-matrix-construction} and Algorithm~\ref{alg:cosep-pruning-procedure} summarize the full procedure.

\subsection{Problem setup and notation}
\label{sec:problem-setup}
Consider a trained network $f(\cdot;\theta)$ and a labeled calibration set
$\mathcal{D}_{\mathrm{cal}}=\{(x_n,y_n)\}_{n=1}^N$, where $y_n \in \{1,\dots,C\}$. Let $\mathcal{L}$ denote the set of prunable layers.
For each layer $\ell \in \mathcal{L}$, let $N_\ell$ be the number of structured components: output channels for convolutional layers and output neurons for linear layers. We write the $c$-th component in layer $\ell$ as $u_{\ell,c}$.

For each input $x$, we extract a scalar activation summary for every component. If layer $\ell$ is convolutional and produces a feature map
$A_\ell(x)\in\mathbb{R}^{N_\ell\times H_\ell \times W_\ell}$, we define
\begin{equation}
z_{\ell,c}(x)=\frac{1}{H_\ell W_\ell}\sum_{h=1}^{H_\ell}\sum_{w=1}^{W_\ell}A_\ell(x)_{c,h,w}.
\label{eq:convolutional-activation-summary}
\end{equation}
If layer $\ell$ is linear and produces a vector
$h_\ell(x)\in\mathbb{R}^{N_\ell}$, we define
\begin{equation}
z_{\ell,c}(x)=h_\ell(x)_c.
\label{eq:linear-activation-summary}
\end{equation}

Thus, each structured component is associated with a class-conditional scalar activation distribution. CoSeP is supervised: labels are used only to estimate these class-conditional activation statistics.

\subsection{Class-separability profiles}
\label{sec:class-separability-profiles}
For each layer $\ell$, component $c$, and class $t$, let
\begin{equation}
\mathcal{Z}_{\ell,c}^{(t)}=\{z_{\ell,c}(x_n)\;|\;y_n=t\}.
\label{eq:class-conditional-activation-set}
\end{equation}
Following the standard closed-form JM construction, we model each
class-conditional activation distribution by a univariate Gaussian,
\begin{equation}
z_{\ell,c}\mid y=t \;\approx\; \mathcal{N}\!\left(\mu_{\ell,c}^{(t)},(\sigma_{\ell,c}^{(t)})^2\right),
\label{eq:class-conditional-gaussian}
\end{equation}
where $\mu_{\ell,c}^{(t)}$ and $\sigma_{\ell,c}^{(t)}$ are the empirical mean and standard deviation computed from $\mathcal{Z}_{\ell,c}^{(t)}$.
This approximation enables a closed-form computation of the Bhattacharyya distance and the corresponding JM separability score from first- and second-order activation statistics.

For the Gaussian model in Equation~\eqref{eq:class-conditional-gaussian}, the closed-form Bhattacharyya distance is~\cite{kailath1967divergence}
\begin{equation}
\mathrm{BD}_{\ell,c}^{(i,j)}
=
\frac{1}{4}\frac{\left(\mu_{\ell,c}^{(i)}-\mu_{\ell,c}^{(j)}\right)^2}
{(\sigma_{\ell,c}^{(i)})^2+(\sigma_{\ell,c}^{(j)})^2+\varepsilon}
+
\frac{1}{2}
\log
\frac{(\sigma_{\ell,c}^{(i)})^2+(\sigma_{\ell,c}^{(j)})^2+\varepsilon}
{2\sigma_{\ell,c}^{(i)}\sigma_{\ell,c}^{(j)}+\varepsilon},
\label{eq:bhattacharyya-distance}
\end{equation}
where $\varepsilon>0$ is a small constant for numerical stability.

We then convert it to a bounded JM separability score:
\begin{equation}
d_{\ell,c}^{(i,j)} = 2\left(1-\exp\left(-\mathrm{BD}_{\ell,c}^{(i,j)}\right)\right).
\label{eq:jm-separability-score}
\end{equation}
Larger values indicate better separation between classes $i$ and $j$ by component $u_{\ell,c}$.

The \emph{class-separability profile} of component $u_{\ell,c}$ is the vector
\begin{equation}
\mathbf{s}_{\ell,c}
=
\left[
d_{\ell,c}^{(i,j)}
\right]_{(i,j)\in\mathcal{P}_\ell}
\in \mathbb{R}^{|\mathcal{P}_\ell|},
\label{eq:class-separability-profile}
\end{equation}
where $\mathcal{P}_\ell$ is the set of class pairs used for layer $\ell$. 
For ImageNet-1K, evaluating all $C(C-1)/2$ class pairs would require 499,500 comparisons per component. We instead rank classes independently in each layer using $C$ one-vs-rest JM scores, retain the top $M$ classes, and form the profile from their $M(M-1)/2$ pairs, reducing pair construction to $O(C+M^2)$. With $M=100$, this yields 4,950 pairs. The sensitivity study in Appendix~\ref{sec:appendix-imagenet-class-count} (Figure~\ref{fig:imagenet-class-count-sensitivity}) shows that accuracy plateaus near $M=100$, which we therefore use for ImageNet-1K.

The set $\{\mathbf{s}_{\ell,c}\}_{c=1}^{N_\ell}$ forms the separability space of layer $\ell$.
Distances in this space are measured with the Euclidean metric:
\begin{equation}
\delta(\mathbf{s}_{\ell,c},\mathbf{s}_{\ell,c'}) =
\|\mathbf{s}_{\ell,c}-\mathbf{s}_{\ell,c'}\|_2.
\label{eq:separability-space-distance}
\end{equation}

\subsection{Complementarity-aware grouping}
\label{sec:complementarity-aware-grouping}
Given the separability profiles of layer $\ell$, CoSeP groups components by applying $k$-medoids clustering in the separability space.
For a candidate number of clusters $k$, the clustering returns a partition
$\{\mathcal{C}_{\ell,1},\dots,\mathcal{C}_{\ell,k}\}$ and corresponding medoids
$\{\mathbf{m}_{\ell,1},\dots,\mathbf{m}_{\ell,k}\}$.

The role of clustering here is not merely compression.
Each cluster collects components with similar class-separability behavior; retaining one representative from each cluster therefore yields a subset that covers multiple discriminative directions rather than repeatedly selecting components that behave similarly.

\subsection{Automatic determination of the retained cardinality}
\label{sec:automatic-retained-cardinality}
The number of retained components should reflect the intrinsic redundancy structure of the layer. To select it automatically, we evaluate the quality of each candidate clustering using the Mean Simplified Silhouette (MSS).

Let $q(c)$ denote the cluster assignment of component $u_{\ell,c}$ under a given $k$.
Define
\begin{equation}
a_{\ell,c} = \delta(\mathbf{s}_{\ell,c}, \mathbf{m}_{\ell,q(c)}),
\label{eq:assigned-medoid-distance}
\end{equation}
that is, the distance from the component to its assigned medoid, and
\begin{equation}
b_{\ell,c} = \min_{r \neq q(c)} \delta(\mathbf{s}_{\ell,c}, \mathbf{m}_{\ell,r}),
\label{eq:nearest-alternative-medoid-distance}
\end{equation}
the distance to the nearest medoid of any other cluster.
The simplified silhouette of component $u_{\ell,c}$ is then
\begin{equation}
\tilde{s}_{\ell,c}
=
\frac{b_{\ell,c}-a_{\ell,c}}
{\max\{a_{\ell,c},\,b_{\ell,c}\}}.
\label{eq:simplified-silhouette}
\end{equation}
The MSS score for layer $\ell$ at cluster count $k$ is
\begin{equation}
\mathrm{MSS}_\ell(k)=\frac{1}{N_\ell}\sum_{c=1}^{N_\ell}\tilde{s}_{\ell,c}.
\label{eq:mean-simplified-silhouette}
\end{equation}

We evaluate $\mathrm{MSS}_\ell(k)$ over a candidate range of cluster counts and obtain a discrete MSS curve. In practice, the curve typically rises quickly for small $k$ and then saturates. We therefore fit a degree-$p$ polynomial to the points $\{(k,\mathrm{MSS}_\ell(k))\}$ to obtain a smoothed curve $\widehat{\mathrm{MSS}}_\ell(k)$, and apply the Kneedle algorithm to this smoothed curve. The detected knee,
\begin{equation}
k_\ell^\star = \mathrm{Kneedle}\!\left(\widehat{\mathrm{MSS}}_\ell(k)\right),
\label{eq:retained-cardinality}
\end{equation}
is taken as the number of components to retain in layer $\ell$. Intuitively, $k_\ell^\star$ is the point beyond which adding more components yields diminishing gains in separability coverage. In all main experiments, we use $p=2$, as justified by the ablation study in Section~\ref{sec:polynomial-degree-ablation}.

\subsection{Representative selection within each cluster}
\label{sec:cluster-representative-selection}
After determining $k_\ell^\star$, we rerun $k$-medoids with that cluster count. Selecting the medoid itself favors geometric centrality, but ignores the strength of the underlying parameters. We therefore retain, from each cluster, the component with the largest parameter norm.

For component $u_{\ell,c}$, define its weight score as
\begin{equation}
w_{\ell,c}=\|W_{\ell,c}\|_2,
\label{eq:component-weight-score}
\end{equation}
where $W_{\ell,c}$ denotes the parameter tensor associated with that component: the output filter for convolutional layers and the corresponding output row for linear layers. The retained representative of cluster $\mathcal{C}_{\ell,r}$ is
\begin{equation}
c_{\ell,r}^\star = \arg\max_{c\in\mathcal{C}_{\ell,r}} w_{\ell,c}.
\label{eq:cluster-representative}
\end{equation}
The retained set for layer $\ell$ is
\begin{equation}
\mathcal{S}_\ell=\{c_{\ell,1}^\star,\dots,c_{\ell,k_\ell^\star}^\star\}.
\label{eq:retained-component-set}
\end{equation}
All components not in $\mathcal{S}_\ell$ are removed.

This selection rule preserves the complementarity induced by the clustering while favoring, within each complementary group, the component with the strongest learned parameters. The representative-selection ablation is reported in Appendix~\ref{sec:representative-selection-ablation}.

\subsection{Layer-wise pruning procedure}
\label{sec:layer-wise-pruning-procedure}
CoSeP is applied sequentially across prunable layers. Let $f^{(0)}=f$ be the original network. For each layer $\ell \in \mathcal{L}$, in network order, we:
\begin{enumerate}
    \item compute component activation summaries on $\mathcal{D}_{\mathrm{cal}}$ using the current network $f^{(\ell-1)}$;
    \item build the separability profiles $\{\mathbf{s}_{\ell,c}\}_{c=1}^{N_\ell}$;
    \item determine $k_\ell^\star$ using MSS and Kneedle;
    \item retain the representative set $\mathcal{S}_\ell$ and prune the remaining components;
    \item briefly fine-tune the pruned network to allow subsequent layers to adapt.
\end{enumerate}
This produces the updated model $f^{(\ell)}$. The exact fine-tuning protocol (optimizer, learning-rate schedule, number of epochs, and calibration/training fraction) is given in Appendix~\ref{sec:appendix-fine-tuning-hyperparameters}.

\subsection{Complexity and practical remarks}
\label{sec:complexity-and-practical-remarks}
For layer $\ell$, profile construction requires computing class-conditional means and variances for $N_\ell$ components over $|\mathcal{P}_\ell|$ class pairs. The dominant selection cost is evaluating $k$-medoids and MSS across candidate $k$ values. When the number of classes is moderate, we use all class pairs. When the class count is large, we use the top-$M$ layer-specific classes described above, and build the profile over all unordered pairs among them, which reduces the profile dimension from $\binom{C}{2}$ to $\binom{M}{2}$. The additional cost of iterative fine-tuning is separate from the subset-selection step. Appendix~\ref{sec:appendix-pruning-overhead} reports detailed pruning times for all models, showing that the full CoSeP pipeline requires 4--12\% of the original training time, with no RL episodes, evolutionary search, or auxiliary-network training.

\section{Experiments}
\label{sec:experiments}

\subsection{Experimental Setup}
\label{sec:experimental-setup}

\paragraph{Datasets and architectures.} We evaluate on CIFAR-10, CIFAR-100, and ImageNet-1K. For CIFAR-10 we test ResNet-56, MobileNet-V2, and VGG-16. For CIFAR-100 we test VGG-16, VGG-19, and DenseNet-40. For ImageNet-1K we test ResNet-50 and MobileNet-V2.

\paragraph{Implementation details.} All experiments use the CoSeP default settings: polynomial degree $P{=}2$ for Kneedle fitting, Weighted component selection, and $M{=}100$ selected classes for ImageNet (see Appendix~\ref{sec:appendix-imagenet-class-count} for the justification of this choice). Fine-tuning after each layer uses 2 epochs on 25\% of the training set for CIFAR-10/100, and 3 epochs on 25\% of the training set for ImageNet-1K, with the same learning rate and optimizer as the original training run. The complete fine-tuning hyperparameters are listed in Appendix~\ref{sec:appendix-fine-tuning-hyperparameters}. All CoSeP and Random Pruning results are averaged over 5 independent runs with different random seeds; we report mean and standard deviation. All experiments were run on 4$\times$ NVIDIA Quadro RTX 6000 GPUs.

\paragraph{Random Pruning baseline.} We include a \emph{Random Pruning} baseline that applies the same per-layer pruning ratios determined by CoSeP but selects which components to remove uniformly at random. Since pruning ratios are identical, the theoretical speedup is the same; accuracy differences isolate the contribution of CoSeP's complementary selection criterion.

\paragraph{Auto column.} The \textit{Auto} column marks methods that automatically determine the per-layer pruning ratio without requiring a user-specified compression target per layer. Methods that apply a uniform global budget distributed manually per layer are not marked.

\subsection{CIFAR-10 Results}
\label{sec:cifar10-results}

\begin{table}[b]
\centering
\small
\caption{
  \textbf{Results on CIFAR-10.} \textit{Base}/\textit{Pruned}: top-1 accuracy (\%); $\Delta$ = Pruned $-$ Base. \textit{Speed Up}: FLOPs ratio before/after pruning. Best and second-best per architecture are \textbf{bold} and \underline{underlined}.
}
\label{tab:cifar10-results}
\begin{minipage}[t]{0.49\linewidth}
\centering
\resizebox{\linewidth}{!}{%
\begin{tabular}{p{4.0cm}ccccc}
\toprule
\textbf{Method} & \textbf{Auto} & \textbf{Base} & \textbf{Pruned} & \textbf{$\Delta$} & \textbf{Speed Up} \\
\midrule
\multicolumn{6}{c}{\textit{--- ResNet-56 ---}} \\
\midrule
DECORE-450 \hfill \cite{alwani2022decore} & \ding{51} & 93.26 & 93.34 & 0.08 & 1.32$\times$ \\
GReg \hfill \cite{he2022filter} &  & 93.51 & 93.28 & -0.23 & 1.99$\times$ \\
AMC \hfill \cite{he2018amc} & \ding{51} & 92.80 & 91.90 & -0.90 & 2.00$\times$ \\
DECORE-200 \hfill \cite{alwani2022decore} & \ding{51} & 93.26 & 93.26 & 0.00 & 2.00$\times$ \\
CP \hfill \cite{li2016pruning} &  & 92.80 & 91.80 & -1.00 & 2.00$\times$ \\
HRank \hfill \cite{lin2020hrank} &  & 93.26 & 93.17 & -0.09 & 2.00$\times$ \\
FPGM \hfill \cite{he2019filter} &  & 93.59 & 93.26 & -0.33 & 2.11$\times$ \\
SNF \hfill \cite{lee2022snf} & \ding{51} & 93.61 & 93.75 & 0.14 & 2.13$\times$ \\
CSPrune \hfill \cite{chen2024csprune} &  & 93.39 & 90.60 & -2.79 & 2.13$\times$ \\
Torque \hfill \cite{gupta2024torque} &  & 93.48 & 93.76 & \textbf{0.28} & 2.15$\times$ \\
AAP \hfill \cite{zhao2023aap} & \ding{51} & 92.84 & 91.78 & -1.06 & 2.17$\times$ \\
Electrostatic-p \hfill \cite{ferdi2024electrostatic} &  & 94.05 & 93.88 & -0.17 & 2.17$\times$ \\
ABCPruner \hfill \cite{lin2020channel} & \ding{51} & 93.26 & 93.23 & -0.03 & 2.18$\times$ \\
ATO \hfill \cite{wu2024auto} & \ding{51} & 93.50 & 93.74 & \underline{0.24} & 2.22$\times$ \\
SCOP \hfill \cite{tang2020scop} &  & 92.84 & 92.90 & 0.06 & 2.27$\times$ \\
REPrune \hfill \cite{park2024reprune} &  & 93.59 & 93.40 & -0.19 & \textbf{2.52$\times$} \\
Random Pruning \hfill &  & 93.58 & 91.75{\scriptsize$\pm$0.31} & -1.83 & \underline{2.28$\times$} \\
\rowcolor{green!10}
CoSeP (ours) & \ding{51} & 93.58 & 93.82{\scriptsize$\pm$0.09} & \underline{0.24} & \underline{2.28$\times$} \\
\bottomrule
\end{tabular}%
}
\end{minipage}\hfill
\begin{minipage}[t]{0.49\linewidth}
\centering
\resizebox{\linewidth}{!}{%
\begin{tabular}{p{4.0cm}ccccc}
\toprule
\textbf{Method} & \textbf{Auto} & \textbf{Base} & \textbf{Pruned} & \textbf{$\Delta$} & \textbf{Speed Up} \\
\midrule
\multicolumn{6}{c}{\textit{--- MobileNet-V2 ---}} \\
\midrule
DMC \hfill \cite{gao2020discrete} &  & 94.23 & 94.49 & \underline{0.26} & 1.66$\times$ \\
SCOP \hfill \cite{tang2020scop} &  & 94.48 & 94.24 & -0.24 & 1.67$\times$ \\
ATO \hfill \cite{wu2024auto} & \ding{51} & 94.45 & 94.78 & \textbf{0.33} & 1.84$\times$ \\
AMC \hfill \cite{he2018amc} & \ding{51} & 94.48 & 93.48 & -1.00 & 1.86$\times$ \\
Random Pruning \hfill &  & 94.48 & 92.51{\scriptsize$\pm$0.34} & -1.97 & 1.93$\times$ \\
AAP \hfill \cite{zhao2023aap} & \ding{51} & 94.46 & 94.70 & 0.24 & \textbf{1.97$\times$} \\
\rowcolor{green!10}
CoSeP (ours) & \ding{51} & 94.48 & 94.81{\scriptsize$\pm$0.08} & \textbf{0.33} & \underline{1.93$\times$} \\
\midrule
\multicolumn{6}{c}{\textit{--- VGG-16 ---}} \\
\midrule
AMC \hfill \cite{he2018amc} & \ding{51} & 93.55 & 92.15 & -1.40 & 1.98$\times$ \\
HRank \hfill \cite{lin2020hrank} &  & 93.64 & 93.11 & -0.53 & 2.15$\times$ \\
GCNP \hfill \cite{jiang2022channel} &  & 93.27 & 93.08 & -0.19 & 2.34$\times$ \\
DECORE-500 \hfill \cite{alwani2022decore} & \ding{51} & 93.96 & 94.02 & 0.06 & 2.35$\times$ \\
CSPrune \hfill \cite{chen2024csprune} &  & 93.75 & 92.57 & -1.18 & 2.55$\times$ \\
AAP \hfill \cite{zhao2023aap} & \ding{51} & 93.64 & 93.82 & \underline{0.18} & \underline{2.57$\times$} \\
Random Pruning \hfill &  & 93.55 & 91.88{\scriptsize$\pm$0.28} & -1.67 & 2.59$\times$ \\
\rowcolor{green!10}
CoSeP (ours) & \ding{51} & 93.55 & 93.92{\scriptsize$\pm$0.07} & \textbf{0.37} & \textbf{2.59$\times$} \\
\bottomrule
\end{tabular}%
}
\end{minipage}
\end{table}

Table~\ref{tab:cifar10-results} reports results on CIFAR-10.

\noindent\textbf{MobileNet-V2.} CoSeP achieves 94.81\% pruned accuracy ($+0.33\%$ gain) at a $1.93\times$ FLOPs reduction, matching ATO's~\cite{wu2024auto} accuracy gain while delivering higher compression. The Random Pruning baseline (92.51\%, same FLOPs reduction) confirms that CoSeP's complementary selection, not merely the pruning ratio, drives the strong result.

\noindent\textbf{VGG-16.} CoSeP reaches 93.92\% ($+0.37\%$) at a $2.59\times$ FLOPs reduction, the largest among the compared methods. AAP~\cite{zhao2023aap} reports $+0.18\%$ at $2.57\times$, while CSPrune~\cite{chen2024csprune} reports $-1.18\%$ at $2.55\times$. Random Pruning at the same ratio yields 91.88\% ($-1.67\%$), further demonstrating the value of the selection criterion.

\noindent\textbf{ResNet-56.} CoSeP attains 93.82\% ($+0.24\%$) at a $2.28\times$ FLOPs reduction; REPrune~\cite{park2024reprune} reaches 93.40\% ($-0.19\%$) at $2.52\times$. These nearby operating points trade accuracy for compression. CoSeP's 2.07-point gap over Random Pruning at matched ratios supports complementarity-aware selection.

\subsection{ImageNet-1K Results}
\label{sec:imagenet-results}

\begin{table}[t]
\centering
\small
\caption{
  \textbf{Results on ImageNet-1K.} \textit{Base}/\textit{Pruned}: top-1 accuracy (\%); $\Delta$ = Pruned $-$ Base. \textit{Speed Up}: FLOPs ratio before/after pruning. Best and second-best per architecture are \textbf{bold} and \underline{underlined}. For CoSeP, ImageNet embeddings use the top-$M$ selected classes with $M=100$.
}
\label{tab:imagenet-results}
\begin{minipage}[t]{0.49\linewidth}
\centering
\resizebox{\linewidth}{!}{%
\begin{tabular}{p{4.0cm}ccccc}
\toprule
\textbf{Method} & \textbf{Auto} & \textbf{Base} & \textbf{Pruned} & \textbf{$\Delta$} & \textbf{Speed Up} \\
\midrule
\multicolumn{6}{c}{\textit{--- ResNet-50 ---}} \\
\midrule
AAP-P \hfill \cite{zhao2023aap} & \ding{51} & 75.06 & 74.42 & -0.64 & 1.55$\times$ \\
GETA \hfill \cite{qu2025geta} & \ding{51} & 76.41 & 76.99 & 0.58 & 2.00$\times$ \\
SMCP \hfill \cite{humble2022soft} &  & 76.20 & 76.80 & 0.60 & 2.15$\times$ \\
DECORE \hfill \cite{alwani2022decore} & \ding{51} & 76.15 & 76.31 & 0.16 & 2.15$\times$ \\
REPrune \hfill \cite{park2024reprune} &  & 76.39 & 77.04 & \underline{0.65} & \underline{2.28$\times$} \\
ATO \hfill \cite{wu2024auto} & \ding{51} & 76.13 & 76.59 & 0.46 & \textbf{2.30$\times$} \\
Random Pruning \hfill &  & 76.32 & 74.58{\scriptsize$\pm$0.22} & -1.74 & \textbf{2.30$\times$} \\
\rowcolor{green!10}
CoSeP (ours) & \ding{51} & 76.32 & 76.98{\scriptsize$\pm$0.11} & \textbf{0.66} & \textbf{2.30$\times$} \\
\bottomrule
\end{tabular}%
}
\end{minipage}\hfill
\begin{minipage}[t]{0.49\linewidth}
\centering
\resizebox{\linewidth}{!}{%
\begin{tabular}{p{4.0cm}ccccc}
\toprule
\textbf{Method} & \textbf{Auto} & \textbf{Base} & \textbf{Pruned} & \textbf{$\Delta$} & \textbf{Speed Up} \\
\midrule
\multicolumn{6}{c}{\textit{--- MobileNet-V2 ---}} \\
\midrule
CC \hfill \cite{li2021towards} &  & 71.88 & 70.91 & -0.97 & 1.39$\times$ \\
SANP \hfill \cite{gao2023sanp} &  & 71.91 & 72.05 & \textbf{0.14} & 1.41$\times$ \\
Uniform \hfill \cite{zhuang2018dcp} &  & 71.80 & 69.80 & -2.00 & 1.42$\times$ \\
AMC \hfill \cite{he2018amc} & \ding{51} & 71.80 & 70.80 & -1.00 & 1.43$\times$ \\
ATO \hfill \cite{wu2024auto} & \ding{51} & 71.88 & 72.02 & \textbf{0.14} & 1.44$\times$ \\
MetaPruning \hfill \cite{liu2019metapruning} & \ding{51} & 72.00 & 71.20 & -0.80 & \underline{1.45$\times$} \\
Random Pruning \hfill &  & 71.90 & 70.09{\scriptsize$\pm$0.26} & -1.81 & \textbf{1.48$\times$} \\
\rowcolor{green!10}
CoSeP (ours) & \ding{51} & 71.90 & 72.01{\scriptsize$\pm$0.10} & \underline{0.11} & \textbf{1.48$\times$} \\
\bottomrule
\end{tabular}%
}
\end{minipage}
\end{table}

Table~\ref{tab:imagenet-results} reports results on ImageNet-1K.

\noindent\textbf{MobileNet-V2.} CoSeP reaches 72.01\% ($+0.11\%$) at a $1.48\times$ FLOPs reduction, the largest reduction among methods with positive~$\Delta$. SANP~\cite{gao2023sanp} and ATO~\cite{wu2024auto} report $+0.14\%$ at $1.41\times$ and $1.44\times$, respectively; CoSeP applies more aggressive compression with competitive pruned accuracy.

\noindent\textbf{ResNet-50.} CoSeP attains 76.98\% ($+0.66\%$) at a $2.30\times$ FLOPs reduction, narrowly the highest $\Delta$ among pure-pruning methods. REPrune~\cite{park2024reprune} reports 77.04\% from a 76.39\% baseline ($+0.65\%$) at $2.28\times$, yielding a comparable accuracy--compression trade-off under its training-integrated regime. GETA~\cite{qu2025geta} combines pruning with quantization to reach 76.99\% ($+0.58\%$) at $2.00\times$; CoSeP applies greater compression while reaching only 0.01 percentage points lower pruned accuracy. The gap between CoSeP and Random Pruning ($-1.74\%$, same FLOPs reduction) confirms that complementarity-aware selection is the key driver.

At the same per-layer ratios, CoSeP exceeds Random Pruning by 2.40 percentage points on ResNet-50/ImageNet-1K, indicating that the selection criterion preserves supervised classification structure better than random selection. This result should not be interpreted as evidence of downstream transfer, which we do not evaluate.
CIFAR-100 results are reported in Appendix~\ref{sec:appendix-cifar100-results}.

\subsection{Ablation Study}
\label{sec:ablation-study}

\subsubsection{Validation of the Gaussian approximation.}
\label{sec:gaussian-approximation-validation}
\begin{figure}[H]
    \centering
    \includegraphics[width=\textwidth]{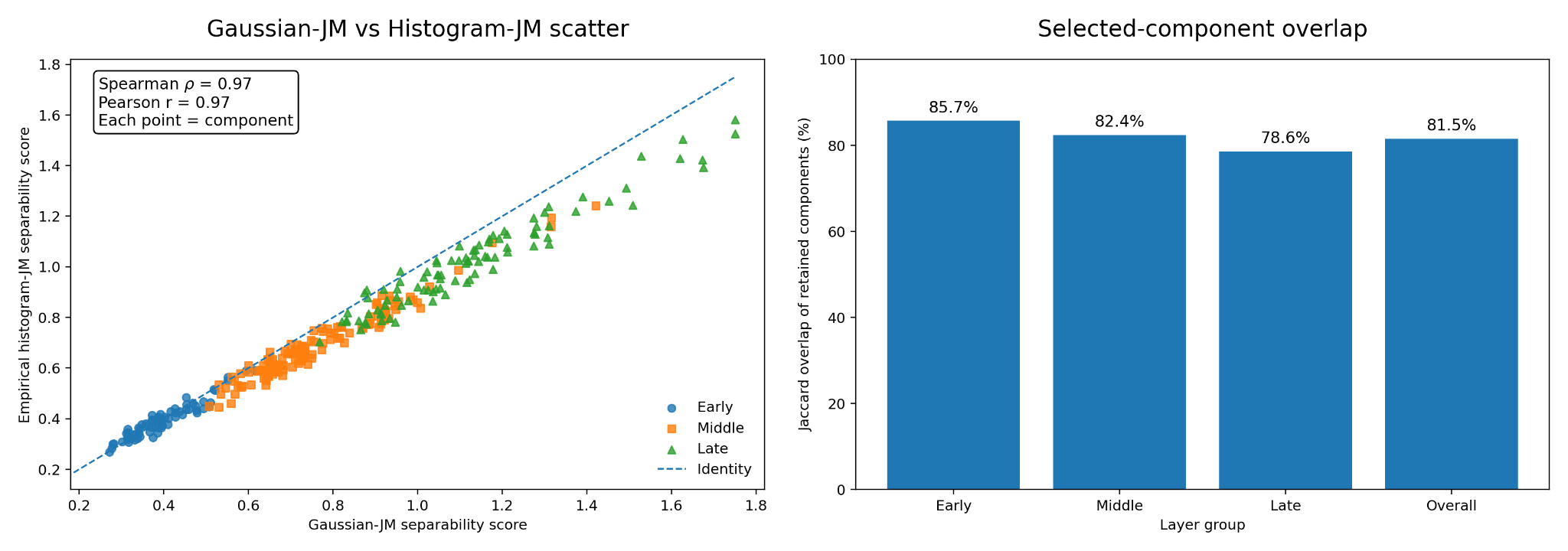}
    \caption{\textbf{Agreement between Gaussian and empirical-histogram JM estimates on VGG-16/CIFAR-10.}
    \textit{Left:} each point is a component; Gaussian- and histogram-based JM scores have Spearman correlation $\rho=0.97$ (dashed line: identity). Colors denote early, middle, and late layers.
    \textit{Right:} Jaccard overlap between retained component sets is 85.7\%, 82.4\%, and 78.6\% by layer group and 81.5\% overall. All other CoSeP settings are fixed.}
    \label{fig:gaussian-histogram-agreement}
\end{figure}

To test whether the Gaussian model in Equation~\eqref{eq:class-conditional-gaussian} materially changes component selection, we recomputed CoSeP with a non-parametric histogram estimate of JM while keeping all other settings fixed. On VGG-16/CIFAR-10, Gaussian- and histogram-based component scores have Spearman correlation $\rho=0.97$, and the corresponding retained sets have 81.5\% Jaccard overlap overall (85.7\%, 82.4\%, and 78.6\% for early, middle, and late layer groups). On ResNet-56/CIFAR-10, $\rho=0.94$ and the retained-set overlap is 73.6\%. Thus, in these two tested settings, the approximation preserves the component ranking and most selected components; we do not claim that individual activation distributions are exactly Gaussian.
Figure~\ref{fig:gaussian-histogram-agreement} visualizes the score agreement and retained-set stability summarized above. Appendix~\ref{sec:appendix-gaussian-approximation} (Figure~\ref{fig:activation-gaussian-fits}) provides the complementary distribution-level diagnostic across network depth.

\subsubsection{Separability Metric Choice}
\label{sec:separability-metric-ablation}

The JM distance was chosen for two properties: (i)~boundedness in $[0,2]$, preventing highly separable class pairs from dominating the embedding geometry, and (ii)~exponential saturation, which keeps extreme values from distorting cluster structure. Table~\ref{tab:separability-metric-comparison} evaluates this empirically by replacing JM with Wasserstein~\cite{ruschendorf1985wasserstein} and Hellinger~\cite{beran1977minimum} distances in the embedding step, keeping all other CoSeP components identical ($P{=}2$, Weighted). JM consistently yields the highest pruned accuracy across all three configurations: on VGG-16/CIFAR-10 it achieves $+0.37\%$ vs.\ $+0.31\%$ (Hellinger) and $+0.25\%$ (Wasserstein); on ResNet-56/CIFAR-10 it achieves $+0.24\%$; on ResNet-50/ImageNet the gap is $+0.66\%$ vs.\ $+0.55\%$ and $+0.49\%$. All three metrics produce positive accuracy changes, indicating that the separability embedding itself is the primary contributor; the JM-specific saturation and boundedness properties provide an additional edge by producing tighter clusters in the embedding space.

\begin{table}[t]
\centering
\small
\caption{
\textbf{Separability metric comparison.}
Effect of replacing the JM distance with Wasserstein or Hellinger distance in the separability embedding, keeping all other components of CoSeP identical ($P{=}2$, Weighted selection). \textbf{Base}: pre-pruning top-1 accuracy (\%); \textbf{Pruned}: post-pruning accuracy; $\Delta$ = Pruned $-$ Base; \textbf{FLOPs\%}: remaining FLOPs after pruning (lower is better). All results averaged over 5 runs (mean{\scriptsize$\pm$std}).
}

\label{tab:separability-metric-comparison}
\resizebox{\columnwidth}{!}{%
\begin{tabular}{llccccc}
\toprule
\textbf{Dataset} & \textbf{Model} & \textbf{Metric} & \textbf{Base Model} & \textbf{Pruned Model} & \textbf{$\Delta$} & \textbf{FLOPs\%} \\
\midrule
\multirow{3}{*}{CIFAR-10} & \multirow{3}{*}{VGG-16}
  & Hellinger~\cite{beran1977minimum}    & 93.55 & 93.86{\scriptsize$\pm$0.10} & +0.31 & 39.05 \\
& & Wasserstein~\cite{ruschendorf1985wasserstein} & 93.55 & 93.80{\scriptsize$\pm$0.12} & +0.25 & 39.72 \\
& & \textbf{JM (ours)}~\cite{kailath1967divergence}   & 93.55 & \textbf{93.92}{\scriptsize$\pm$0.07} & \textbf{+0.37} & \textbf{38.61} \\
\midrule
\multirow{3}{*}{CIFAR-10} & \multirow{3}{*}{ResNet-56}
  & Hellinger~\cite{beran1977minimum}    & 93.58 & 93.76{\scriptsize$\pm$0.11} & +0.18 & 44.32 \\
& & Wasserstein~\cite{ruschendorf1985wasserstein} & 93.58 & 93.72{\scriptsize$\pm$0.13} & +0.14 & 44.70 \\
& & \textbf{JM (ours)}~\cite{kailath1967divergence}   & 93.58 & \textbf{93.82}{\scriptsize$\pm$0.09} & \textbf{+0.24} & \textbf{43.86} \\
\midrule
\multirow{3}{*}{ImageNet-1K} & \multirow{3}{*}{ResNet-50}
  & Hellinger~\cite{beran1977minimum}    & 76.32 & 76.87{\scriptsize$\pm$0.08} & +0.55 & 44.02 \\
& & Wasserstein~\cite{ruschendorf1985wasserstein} & 76.32 & 76.81{\scriptsize$\pm$0.10} & +0.49 & 44.38 \\
& & \textbf{JM (ours)}~\cite{kailath1967divergence}   & 76.32 & \textbf{76.98}{\scriptsize$\pm$0.11} & \textbf{+0.66} & \textbf{43.48} \\
\bottomrule
\end{tabular}%
}
\end{table}

\subsubsection{Polynomial Degree}
\label{sec:polynomial-degree-ablation}

Table~\ref{tab:polynomial-representative-ablation} compares representative-selection strategies at polynomial degrees 2--5.

\begin{table*}[ht!]
\centering
\caption{\textbf{Ablation on representative selection and polynomial degree}. For each polynomial degree $P \in \{2,3,4,5\}$. The three variants differ only in how one representative is chosen from each final cluster: \textbf{Random-in-cluster} \textit{(Random)}, \textbf{Medoid} \textit{(Regular)}, or \textbf{Highest-weight} \textit{(Weighted)}. 
We report top-1 accuracy and remaining FLOPs.}
\label{tab:polynomial-representative-ablation}
\resizebox{\textwidth}{!}{
\begin{tabular}{l l l|c c|c c|c c|c c|c c}
\toprule
& & & \multicolumn{2}{c|}{\textbf{Base Model}} & \multicolumn{2}{c|}{\textbf{Poly 2}} & \multicolumn{2}{c|}{\textbf{Poly 3}} & \multicolumn{2}{c|}{\textbf{Poly 4}} & \multicolumn{2}{c}{\textbf{Poly 5}} \\
\midrule
\textbf{Dataset} & \textbf{Model} & \textbf{Method} & \textbf{Accuracy} & \textbf{FLOPs} & \textbf{Accuracy} & \textbf{FLOPs} & \textbf{Accuracy} & \textbf{FLOPs} & \textbf{Accuracy} & \textbf{FLOPs} & \textbf{Accuracy} & \textbf{FLOPs} \\
\midrule

\multirow{9}{*}{\textbf{CIFAR-10}}
& \multirow{3}{*}{MobileNet-V2} & Random
& \multirow{3}{*}{94.48}
& \multirow{3}{*}{\shortstack{$\uparrow$ \\ 100\% \\ $\downarrow$}}
& 92.51
& \multirow{3}{*}{\shortstack{$\uparrow$ \\ 51.81\% \\ $\downarrow$}}
& 88.29
& \multirow{3}{*}{\shortstack{$\uparrow$ \\ 39.62\% \\ $\downarrow$}}
& 85.80
& \multirow{3}{*}{\shortstack{$\uparrow$ \\ 30.80\% \\ $\downarrow$}}
& 84.10
& \multirow{3}{*}{\shortstack{$\uparrow$ \\ 24.07\% \\ $\downarrow$}} \\
& & Regular
& & & 94.51 & & 90.12 & & 87.87 & & 86.34 & \\
& & Weighted
& & & \textbf{94.81} & & \textbf{90.71} & & \textbf{88.82} & & \textbf{87.95} & \\

\cmidrule{2-13}

& \multirow{3}{*}{VGG-16} & Random
& \multirow{3}{*}{93.55}
& \multirow{3}{*}{\shortstack{$\uparrow$ \\ 100\% \\ $\downarrow$}}
& 91.88
& \multirow{3}{*}{\shortstack{$\uparrow$ \\ 38.61\% \\ $\downarrow$}}
& 87.53
& \multirow{3}{*}{\shortstack{$\uparrow$ \\ 29.28\% \\ $\downarrow$}}
& 84.82
& \multirow{3}{*}{\shortstack{$\uparrow$ \\ 22.89\% \\ $\downarrow$}}
& 82.95
& \multirow{3}{*}{\shortstack{$\uparrow$ \\ 17.76\% \\ $\downarrow$}} \\
& & Regular
& & & 93.55 & & 89.44 & & 86.98 & & 85.33 & \\
& & Weighted
& & & \textbf{93.92} & & \textbf{90.01} & & \textbf{87.83} & & \textbf{86.97} & \\

\cmidrule{2-13}

& \multirow{3}{*}{ResNet-56} & Random
& \multirow{3}{*}{93.58}
& \multirow{3}{*}{\shortstack{$\uparrow$ \\ 100\% \\ $\downarrow$}}
& 91.75
& \multirow{3}{*}{\shortstack{$\uparrow$ \\ 43.86\% \\ $\downarrow$}}
& 87.40
& \multirow{3}{*}{\shortstack{$\uparrow$ \\ 35.40\% \\ $\downarrow$}}
& 84.82
& \multirow{3}{*}{\shortstack{$\uparrow$ \\ 27.66\% \\ $\downarrow$}}
& 82.74
& \multirow{3}{*}{\shortstack{$\uparrow$ \\ 23.50\% \\ $\downarrow$}} \\
& & Regular
& & & 93.41 & & 89.28 & & 86.93 & & 85.03 & \\
& & Weighted
& & & \textbf{93.82} & & \textbf{90.21} & & \textbf{87.97} & & \textbf{86.89} & \\

\midrule \midrule

\multirow{9}{*}{\textbf{CIFAR-100}}
& \multirow{3}{*}{VGG-16} & Random
& \multirow{3}{*}{73.70}
& \multirow{3}{*}{\shortstack{$\uparrow$ \\ 100\% \\ $\downarrow$}}
& 72.16
& \multirow{3}{*}{\shortstack{$\uparrow$ \\ 49.75\% \\ $\downarrow$}}
& 68.66
& \multirow{3}{*}{\shortstack{$\uparrow$ \\ 36.14\% \\ $\downarrow$}}
& 66.40
& \multirow{3}{*}{\shortstack{$\uparrow$ \\ 30.38\% \\ $\downarrow$}}
& 65.16
& \multirow{3}{*}{\shortstack{$\uparrow$ \\ 24.10\% \\ $\downarrow$}} \\
& & Regular
& & & 73.95 & & 70.69 & & 68.67 & & 67.57 & \\
& & Weighted
& & & \textbf{74.31} & & \textbf{71.08} & & \textbf{69.68} & & \textbf{68.79} & \\

\cmidrule{2-13}

& \multirow{3}{*}{VGG-19} & Random
& \multirow{3}{*}{73.38}
& \multirow{3}{*}{\shortstack{$\uparrow$ \\ 100\% \\ $\downarrow$}}
& 71.52
& \multirow{3}{*}{\shortstack{$\uparrow$ \\ 47.39\% \\ $\downarrow$}}
& 67.86
& \multirow{3}{*}{\shortstack{$\uparrow$ \\ 38.91\% \\ $\downarrow$}}
& 65.98
& \multirow{3}{*}{\shortstack{$\uparrow$ \\ 33.63\% \\ $\downarrow$}}
& 64.68
& \multirow{3}{*}{\shortstack{$\uparrow$ \\ 27.34\% \\ $\downarrow$}} \\
& & Regular
& & & 73.36 & & 69.95 & & 68.30 & & 67.16 & \\
& & Weighted
& & & \textbf{73.90} & & \textbf{70.89} & & \textbf{68.83} & & \textbf{68.00} & \\

\cmidrule{2-13}

& \multirow{3}{*}{DenseNet-40} & Random
& \multirow{3}{*}{74.30}
& \multirow{3}{*}{\shortstack{$\uparrow$ \\ 100\% \\ $\downarrow$}}
& 71.65
& \multirow{3}{*}{\shortstack{$\uparrow$ \\ 52.35\% \\ $\downarrow$}}
& 68.09
& \multirow{3}{*}{\shortstack{$\uparrow$ \\ 39.62\% \\ $\downarrow$}}
& 66.77
& \multirow{3}{*}{\shortstack{$\uparrow$ \\ 30.84\% \\ $\downarrow$}}
& 65.38
& \multirow{3}{*}{\shortstack{$\uparrow$ \\ 23.98\% \\ $\downarrow$}} \\
& & Regular
& & & 73.51 & & 70.23 & & 69.13 & & 67.91 & \\
& & Weighted
& & & \textbf{73.94} & & \textbf{70.85} & & \textbf{69.34} & & \textbf{68.75} & \\

\midrule \midrule

\multirow{6}{*}{\textbf{ImageNet-1K}}
& \multirow{3}{*}{MobileNet-V2} & Random
& \multirow{3}{*}{71.90}
& \multirow{3}{*}{\shortstack{$\uparrow$ \\ 100\% \\ $\downarrow$}}
& 70.09
& \multirow{3}{*}{\shortstack{$\uparrow$ \\ 67.57\% \\ $\downarrow$}}
& 66.53
& \multirow{3}{*}{\shortstack{$\uparrow$ \\ 49.33\% \\ $\downarrow$}}
& 64.33
& \multirow{3}{*}{\shortstack{$\uparrow$ \\ 38.63\% \\ $\downarrow$}}
& 63.10
& \multirow{3}{*}{\shortstack{$\uparrow$ \\ 32.09\% \\ $\downarrow$}} \\
& & Regular
& & & 71.83 & & 68.46 & & 66.51 & & 65.54 & \\
& & Weighted
& & & \textbf{72.01} & & \textbf{69.19} & & \textbf{67.34} & & \textbf{66.41} & \\

\cmidrule{2-13}

& \multirow{3}{*}{ResNet-50} & Random
& \multirow{3}{*}{76.32}
& \multirow{3}{*}{\shortstack{$\uparrow$ \\ 100\% \\ $\downarrow$}}
& 74.58
& \multirow{3}{*}{\shortstack{$\uparrow$ \\ 43.48\% \\ $\downarrow$}}
& 71.32
& \multirow{3}{*}{\shortstack{$\uparrow$ \\ 36.77\% \\ $\downarrow$}}
& 69.14
& \multirow{3}{*}{\shortstack{$\uparrow$ \\ 28.51\% \\ $\downarrow$}}
& 67.04
& \multirow{3}{*}{\shortstack{$\uparrow$ \\ 24.56\% \\ $\downarrow$}} \\
& & Regular
& & & 76.40 & & 73.29 & & 71.36 & & 69.61 & \\
& & Weighted
& & & \textbf{76.98} & & \textbf{73.98} & & \textbf{72.02} & & \textbf{71.15} & \\

\bottomrule
\end{tabular}
}
\end{table*}

Higher polynomial degrees cause Kneedle to fit a tighter curve to the MSS profile, identifying a smaller~$k^*$ and therefore applying more aggressive pruning. While this reduces FLOPs, it consistently reduces accuracy across all architectures and datasets. Degree~$P{=}2$ provides the best accuracy--compression balance and is adopted as the default. Crucially, once $P$ is fixed, CoSeP requires \emph{no further per-layer manual tuning}, all per-layer compression rates are determined automatically by the MSS knee.

\subsubsection{Effect of the Recovery Schedule}
\label{sec:recovery-schedule-ablation}
Table~\ref{tab:recovery-schedule-ablation} isolates recovery timing under a matched optimization budget. Final-only CoSeP exceeds Random Pruning on both architectures, while per-layer recovery performs best; the remaining gain indicates that refreshing activation statistics after each pruning step is beneficial.

\begin{table}[H]
\centering
\small
\caption{\textbf{Effect of the recovery schedule on CIFAR-10 top-1 accuracy (\%).}
\textit{Final-only} applies the same total recovery budget once after all layers are pruned; \textit{Per-layer} is the default CoSeP schedule; \textit{Random} uses CoSeP's per-layer ratios and the same total recovery budget. Best result per architecture is bold. Both schedules use the same cumulative number of SGD updates, 25\% of the CIFAR-10 training data, and the architecture-specific optimizer settings in Appendix~\ref{sec:appendix-fine-tuning-hyperparameters}, Table~\ref{tab:fine-tuning-hyperparameters}; optimizer-state handling and the learning-rate schedule are held fixed. Per-layer recovery uses two epochs after each pruning step, for \(2|\mathcal{L}|\) recovery epochs in total, whereas final-only recovery applies the matched cumulative update budget once after all layers are pruned.}
\label{tab:recovery-schedule-ablation}
\begin{tabular}{lccc}
\toprule
Model & Final-only & Per-layer & Random \\
\midrule
ResNet-56 & 92.93 & \textbf{93.82} & 91.75 \\
VGG-16 & 92.37 & \textbf{93.92} & 91.88 \\
\bottomrule
\end{tabular}
\end{table}

\subsection{Analysis}
\label{sec:analysis}

\paragraph{Per-layer retention.} A key feature of CoSeP is that it automatically determines a different pruning ratio at each layer, reflecting genuine differences in per-layer redundancy. The resulting FLOPs percentages for the default ($P{=}2$) setting range from 43\% (ResNet-50) to 68\% (MobileNet-V2/ImageNet). Empirically, deeper convolutional layers tend to exhibit higher inter-channel redundancy in the separability space, so CoSeP typically prunes them more aggressively, while retaining a larger fraction of components in earlier layers, consistent with prior observations on layer-wise sensitivity~\cite{blalock2020state,fang2023depgraph}. Appendix~\ref{sec:appendix-per-layer-retention} provides per-layer retention plots and analysis.

\paragraph{Why pruning sometimes improves accuracy.} Several entries show positive~$\Delta$ after pruning. We attribute this to two complementary mechanisms. \emph{(i)~Regularization effect}: removing redundant or noisy components reduces over-fitting, particularly in over-parameterized architectures. \emph{(ii)~Diversity effect}: retaining components with diverse class-separability profiles produces a pruned model whose remaining channels are more independently informative, reducing internal covariate redundancy. Brief post-pruning fine-tuning further allows the network to adapt to the pruned topology. Similar accuracy improvements after pruning have been reported by SCOP~\cite{tang2020scop}, ATO~\cite{wu2024auto}, and Torque~\cite{gupta2024torque}.

\subsection{Inference Time}
\label{sec:inference-time}

\begin{table}[t]
\centering
\caption{
\textbf{Batch and single-image inference times before and after CoSeP pruning.} $\Delta\%$ is the percentage change in wall-clock time. Values are averaged over 100 runs × 5 seeds on 4× NVIDIA Quadro RTX 6000 GPUs (mean$\pm$std across seeds).
}
\label{tab:inference-times}
\resizebox{\columnwidth}{!}{%
\begin{tabular}{llccccccc}
\toprule
\multirow{2}{*}{Dataset} & \multirow{2}{*}{Model} &
\multirow{2}{*}{FLOPs$\times$} &
\multicolumn{3}{c}{Batch Inference (ms)} &
\multicolumn{3}{c}{Single Inference (ms)} \\
\cmidrule(lr){4-6}\cmidrule(lr){7-9}
& & & Full Model & Pruned Model & $\Delta\%$ & Full Model & Pruned Model & $\Delta\%$ \\
\midrule
\multirow{3}{*}{CIFAR-10}
& MobileNet-V2 & $1.93\times$ & 5.339 & 4.249{\scriptsize$\pm$0.04} & $-20.39$ & 3.785 & 3.587{\scriptsize$\pm$0.03} & $-5.23$ \\
& VGG-16       & $2.59\times$ & 1.091 & 0.975{\scriptsize$\pm$0.02} & $-10.63$ & 0.771 & 0.718{\scriptsize$\pm$0.01} & $-6.88$ \\
& ResNet-56    & $2.28\times$ & 4.431 & 4.158{\scriptsize$\pm$0.03} & $-6.16$  & 3.995 & 3.763{\scriptsize$\pm$0.02} & $-5.82$ \\
\midrule
\multirow{3}{*}{CIFAR-100}
& VGG-16       & $2.01\times$ & 0.979 & 0.915{\scriptsize$\pm$0.01} & $-6.54$  & 0.794 & 0.746{\scriptsize$\pm$0.01} & $-6.05$ \\
& VGG-19       & $2.11\times$ & 1.114 & 1.007{\scriptsize$\pm$0.02} & $-9.61$  & 0.938 & 0.881{\scriptsize$\pm$0.01} & $-6.08$ \\
& DenseNet-40  & $1.91\times$ & 4.425 & 4.186{\scriptsize$\pm$0.03} & $-5.40$  & 3.924 & 3.689{\scriptsize$\pm$0.02} & $-5.99$ \\
\midrule
\multirow{2}{*}{ImageNet}
& MobileNet-V2 & $1.48\times$ & 7.636 & 6.814{\scriptsize$\pm$0.05} & $-10.76$ & 6.203 & 5.861{\scriptsize$\pm$0.04} & $-5.51$ \\
& ResNet-50    & $2.30\times$ & 5.255 & 4.923{\scriptsize$\pm$0.04} & $-6.32$  & 4.616 & 4.244{\scriptsize$\pm$0.03} & $-8.07$ \\
\midrule
\multicolumn{2}{l}{\textit{Average}} & & & & $-9.48$ & & & $-6.20$ \\
\bottomrule
\end{tabular}%
}
\end{table}

Table~\ref{tab:inference-times} connects theoretical FLOPs reductions to observed latency for batch and single-image inference.
CoSeP delivers consistent wall-clock reductions across all tested models. For CIFAR-10, MobileNet-V2 achieves the largest batch speedup ($-20.39\%$). For ImageNet, MobileNet-V2 achieves $-10.76\%$ batch and ResNet-50 $-8.07\%$ single-image latency improvement. Averaged across all models, batch inference improves by $9.48\%$ and single-image by $6.20\%$.

\paragraph{FLOPs vs.\ wall-clock speedup.} The $2.30\times$ FLOPs reduction for ResNet-50 does not map linearly to $2.30\times$ wall-clock speedup for two reasons: \emph{(i)~Memory bandwidth bottleneck}, for small batch sizes, GPU kernels are bandwidth-limited rather than compute-limited; fewer channels reduce memory traffic but per-kernel launch overhead remains constant. \emph{(ii)~SIMD underutilisation}, tensor cores operate on fixed-width tiles; pruned layers with channel counts below tile-width boundaries do not fully utilise available parallelism. Despite this, the $5$--$20\%$ wall-clock reductions are practically significant and consistent, confirming that structured pruning delivers real deployment benefits.

\section{Conclusion}
\label{sec:conclusion}

We presented CoSeP (Complementary Separability Pruning), a structured pruning method that selects components by clustering their pairwise class-separability profiles in a separability embedding space. The core novelty is the \emph{complementarity principle}: by applying $k$-medoids over JM-based separability vectors, CoSeP retains components from diverse regions of this space, avoiding the redundancy that plagues magnitude-only and correlation-based criteria. An MSS-guided knee-detection algorithm automatically determines the per-layer retention count, eliminating the need for manual compression ratio specification.

Across CIFAR-10, CIFAR-100, and ImageNet-1K with six architectures, CoSeP consistently matches or surpasses 15+ state-of-the-art methods. On ResNet-50/ImageNet, $+0.66\%$ accuracy at $2.30\times$ FLOPs reduction is the highest $\Delta$ among all pure pruning methods. Hardware inference timing confirms real-world deployment benefits. The consistent gap between CoSeP and Random Pruning at identical per-layer ratios supports the contribution of complementarity-aware selection relative to random selection.

\paragraph{Scope and limitations.}
Our experiments establish CoSeP only for supervised CNN image classification. We have not evaluated transfer to detection or segmentation, self-supervised representations, vision-language or language models, or Transformer architectures; ImageNet top-1 accuracy alone is insufficient evidence for those regimes. A possible Transformer extension would construct class-conditional descriptors for attention heads and replace the static parameter-norm representative score with a head-salience measure, but this hypothesis requires separate methodological and empirical validation.

\newpage
\bibliography{egbib}

\newpage
\appendix

\section{Choice of $M$ for ImageNet-1K}
\label{sec:appendix-imagenet-class-count}

ImageNet-1K has $C=1000$ classes, so using all unordered class pairs would require $\binom{1000}{2}=499{,}500$ pairwise comparisons per layer. Computing the full separability profile at that scale is prohibitively expensive. We therefore first score each class using a layer-specific one-vs-rest JM screening criterion, average that score across components, and retain the top-\(M\) classes. The embedding is then constructed from all unordered pairs among those selected classes, i.e.\ \(\binom{M}{2}\) pairs.

Specifically, for each class \(t\), we pool all samples with labels \(y\neq t\) into a complement group and compute, for each component, a one-vs-rest JM separability score \(d^{(t,\neg t)}_{\ell,c}\) using the same Gaussian/JM construction as in Equations~\eqref{eq:class-conditional-gaussian}--\eqref{eq:jm-separability-score}. We then score class \(t\) by averaging over components:
\begin{equation}
\bar d^{(t)}_\ell =
\frac{1}{N_\ell}
\sum_{c=1}^{N_\ell}
d^{(t,\neg t)}_{\ell,c}.
\label{eq:class-screening-score}
\end{equation}
We define \(T_\ell\) as the set of the \(M\) classes with the largest values of \(\bar d^{(t)}_\ell\), and set
\begin{equation}
\mathcal{P}_\ell = \{(i,j)\mid i,j\in T_\ell,\ i<j\}.
\label{eq:selected-class-pair-set}
\end{equation}

\begin{figure}[H]
    \centering
    \includegraphics[width=\textwidth]{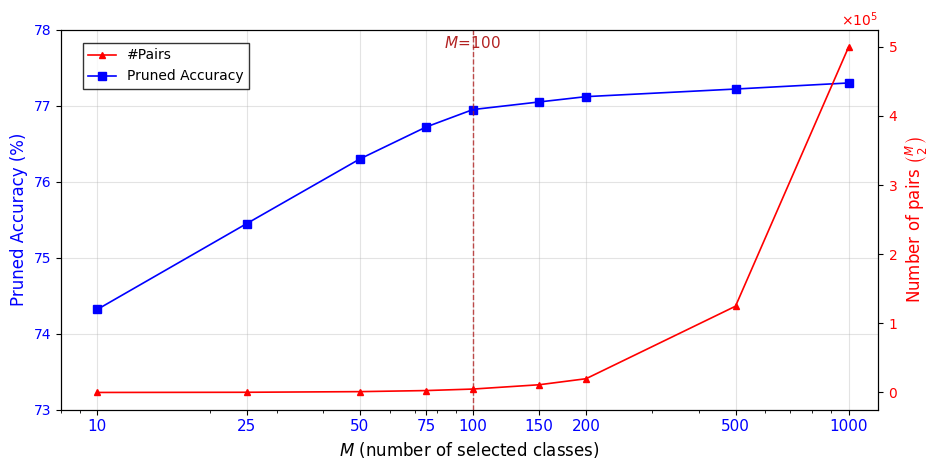}
    \caption{\textbf{Effect of selected-class count $M$ on ResNet-50/ImageNet-1K.}
    Blue: pruned top-1 accuracy for CoSeP with $p=2$ and weighted representative selection. Red: the number of evaluated class pairs, $\binom{M}{2}$. Accuracy changes little beyond $M=100$, whereas the pair count grows from 4,950 at $M=100$ to 499,500 at $M=1000$; the dashed line marks the adopted $M=100$.}
    \label{fig:imagenet-class-count-sensitivity}
\end{figure}

Figure~\ref{fig:imagenet-class-count-sensitivity} evaluates the trade-off between class-pair coverage and computation. The onset of the accuracy plateau near $M=100$ motivates the adopted default without repeating the full pairwise cost of larger values.

\section{Selection Strategy}
\label{sec:representative-selection-ablation}

In all variants, CoSeP keeps the same pipeline for determining the retained cardinality: MSS is evaluated across candidate cluster counts, Kneedle selects $k^*$, and k-medoids is run with $k = k^*$. The ablation changes only the rule used to choose one representative from each final cluster: random-in-cluster, medoid, or highest-weight.

Across all polynomial degrees, architectures, and datasets, \textbf{Weighted} selection consistently outperforms \textbf{Regular} (medoid) selection, which in turn outperforms \textbf{Random-in-cluster} selection.
Since all three variants use the same CoSeP clustering and the same retained cardinality $k^*$, these gaps isolate the effect of the \textbf{within-cluster representative choice}.
The advantage of Weighted over Regular shows that parameter magnitude provides useful information beyond geometric centrality alone.
The gap between Regular and Random-in-cluster further shows that, once the complementary cluster structure is fixed, selecting a principled representative is preferable to choosing an arbitrary cluster member.

\begin{figure}[H]
\centering
\includegraphics[width=0.7\columnwidth]{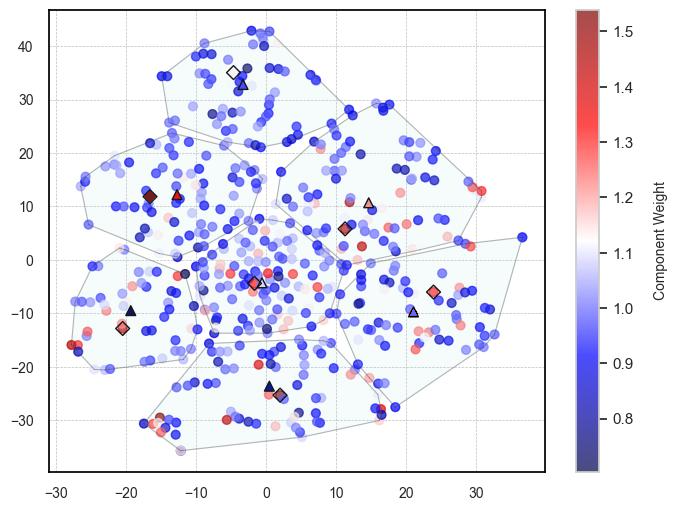}
    \caption{\textbf{Representative selection in the separability space.} This 2-D projection shows a ResNet-56 linear layer's components, colored by weight. The space forms 7 clusters; medoids are triangles, and the highest-weight component in each cluster is a rhombus.}
\label{fig:representative-selection-projection}
\end{figure}

Figure~\ref{fig:representative-selection-projection} illustrates the distinction between geometric centrality and weight-aware selection: a weighted representative can differ from the medoid while preserving one representative per cluster~\cite{vandermaaten2008tsne}.

\section{Fine-Tuning Hyperparameters}
\label{sec:appendix-fine-tuning-hyperparameters}

Table~\ref{tab:fine-tuning-hyperparameters} lists the fine-tuning hyperparameters
used after each layer-wise pruning step. The optimizer, learning rate, and weight decay follow the settings of the original pre-training recipe for each architecture. The number of fine-tuning epochs and the fraction of training data used are kept deliberately small to minimize overhead while allowing the surviving channels to adapt.

\begin{table}[H]
\centering
\caption{\textbf{Fine-tuning hyperparameters per dataset and architecture.} "Epochs" refers to the number of fine-tuning epochs \emph{per pruned layer}; "Data\%" is the fraction of data used.}
\label{tab:fine-tuning-hyperparameters}
\resizebox{\columnwidth}{!}{%
\begin{tabular}{llcccccccc}
\toprule
\textbf{Dataset} & \textbf{Model} & \textbf{Optimizer} & \textbf{LR} & \textbf{Momentum} & \textbf{Weight Decay} & \textbf{Nesterov} & \textbf{Batch Size} & \textbf{Epochs} & \textbf{Data\%} \\
\midrule
\multirow{3}{*}{CIFAR-10}
& ResNet-56    & SGD & 0.01  & 0.9 & $1\!\times\!10^{-4}$ & False & 128 & 2 & 25\% \\
& VGG-16       & SGD & 0.01  & 0.9 & $5\!\times\!10^{-4}$ & True  & 128 & 2 & 25\% \\
& MobileNet-V2 & SGD & 0.01  & 0.9 & $5\!\times\!10^{-4}$ & False & 128 & 2 & 25\% \\
\midrule
\multirow{3}{*}{CIFAR-100}
& VGG-16       & SGD & 0.01  & 0.9 & $5\!\times\!10^{-4}$ & True  & 128 & 2 & 25\% \\
& VGG-19       & SGD & 0.01  & 0.9 & $5\!\times\!10^{-4}$ & True  & 128 & 2 & 25\% \\
& DenseNet-40  & SGD & 0.01  & 0.9 & $5\!\times\!10^{-4}$ & False & 128 & 2 & 25\% \\
\midrule
\multirow{2}{*}{ImageNet-1K}
& ResNet-50    & SGD & 0.001 & 0.9 & $1\!\times\!10^{-4}$ & False & 256 & 3 & 25\% \\
& MobileNet-V2 & SGD & 0.001 & 0.9 & $5\!\times\!10^{-4}$ & False & 256 & 3 & 25\% \\
\bottomrule
\end{tabular}%
}
\end{table}

\section{Pruning Overhead}
\label{sec:appendix-pruning-overhead}

Table~\ref{tab:pruning-overhead} reports the wall-clock time for the full CoSeP pruning pipeline (profile construction, $k$-medoids clustering, and layer-wise fine-tuning) for each model. The dominant cost is the iterative fine-tuning; profile construction and clustering together account for a modest share of the total pruning time, remaining below 20\% across all reported models. Crucially, CoSeP requires no RL episodes, evolutionary search, or controller-network training; its overhead is a single sequential pass through the prunable layers. As a fraction of the original full training time, the total pruning overhead ranges from 4\% to 12\%, making CoSeP substantially cheaper than search-based methods such as AMC~\cite{he2018amc} or ABCPruner~\cite{lin2020channel}.

\begin{table}[H]
\centering
\small
\caption{\textbf{CoSeP pruning overhead.} \textbf{Profile+Cluster}: time for separability profile construction and $k$-medoids clustering (all layers combined). \textbf{Fine-tune}: cumulative layer-wise fine-tuning time. \textbf{Total}: end-to-end pruning time. \textbf{Ratio}: total pruning time as a fraction of original full training time.}
\label{tab:pruning-overhead}
\resizebox{\columnwidth}{!}{%
\begin{tabular}{llcccc}
\toprule
\textbf{Dataset} & \textbf{Model} & \textbf{Profile+Cluster} & \textbf{Fine-tune} & \textbf{Total} & \textbf{Ratio (\% of Training)} \\
\midrule
\multirow{3}{*}{CIFAR-10}
& MobileNet-V2 & 2.4 min  & 18.1 min & 20.5 min  & 5.1\% \\
& VGG-16       & 1.8 min  & 14.6 min & 16.4 min  & 4.3\% \\
& ResNet-56    & 3.1 min  & 22.7 min & 25.8 min  & 6.8\% \\
\midrule
\multirow{3}{*}{CIFAR-100}
& VGG-16       & 4.2 min  & 19.3 min & 23.5 min  & 5.9\% \\
& VGG-19       & 5.1 min  & 24.8 min & 29.9 min  & 6.7\% \\
& DenseNet-40  & 4.8 min  & 21.2 min & 26.0 min  & 7.2\% \\
\midrule
\multirow{2}{*}{ImageNet-1K}
& MobileNet-V2 & 12.3 min & 67.4 min & 79.7 min  & 8.4\% \\
& ResNet-50    & 18.6 min & 98.2 min & 116.8 min & 11.7\% \\
\bottomrule
\end{tabular}%
}
\end{table}

\section{CIFAR-100 Results}
\label{sec:appendix-cifar100-results}

Table~\ref{tab:polynomial-representative-ablation} reports the underlying CIFAR-100 measurements; the values below correspond to the $p=2$ weighted-selection rows.

On VGG-16, CoSeP achieves $73.70{\to}74.31\%$ ($+0.61\%$) at 49.75\% remaining FLOPs, and on VGG-19 $73.38{\to}73.90\%$ ($+0.52\%$) at 47.39\%. DenseNet-40 yields $74.30{\to}73.94\%$ ($-0.36\%$) at 52.35\%, the dense skip-connection topology reduces apparent per-channel redundancy, making compression more challenging.

The Weighted–Regular–Random-in-cluster accuracy ordering holds across all three architectures, confirming that, once CoSeP's cluster structure is fixed, highest-weight representatives are preferable to medoids, and medoids are preferable to arbitrary within-cluster choices.

\section{Gaussian Approximation Validation}
\label{sec:appendix-gaussian-approximation}

\begin{figure}[H]
    \centering
    \includegraphics[width=\textwidth]{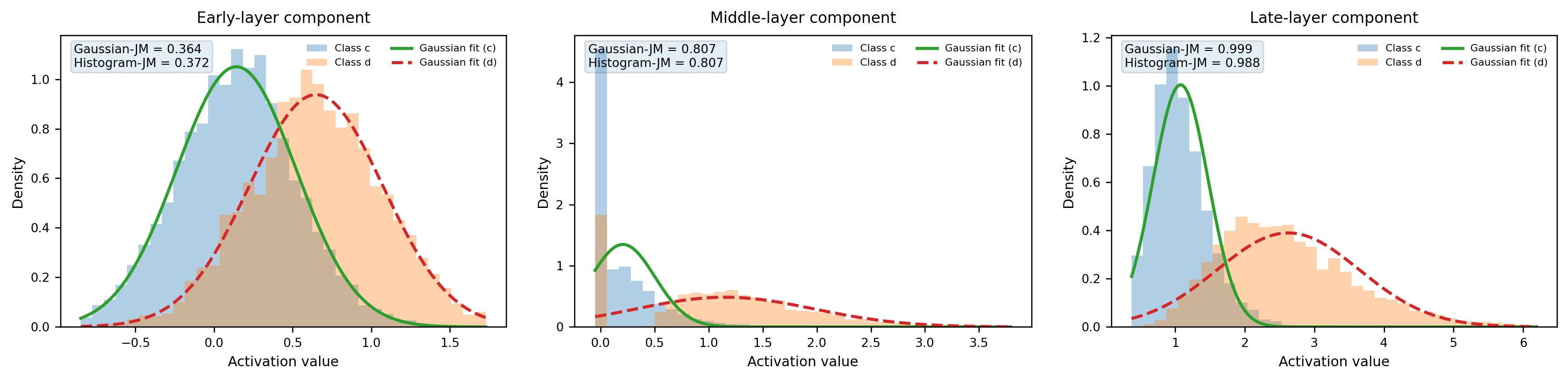}
    \caption{\textbf{Representative class-conditional activation distributions and univariate Gaussian fits.}
    From left to right, panels show components from early, middle, and late layers. Histograms show empirical densities for the two classes, and curves show the fitted Gaussians; insets report Gaussian- and histogram-based JM scores. The middle and late examples visibly depart from an exact Gaussian shape, yet the two JM estimates remain close.}
    \label{fig:activation-gaussian-fits}
\end{figure}

This appendix complements the quantitative stability analysis in Section~\ref{sec:gaussian-approximation-validation} by examining the activation-distribution shapes directly. Figure~\ref{fig:activation-gaussian-fits} shows representative early-, middle-, and late-layer components. Although the middle and late examples visibly depart from an exact Gaussian shape, their Gaussian- and histogram-based JM values remain close. This supports using the Gaussian model as an efficient separability surrogate without treating it as an exact distributional description.

\section{Per-Layer Retention}
\label{sec:appendix-per-layer-retention}

\begin{figure}[H]
    \centering
    \includegraphics[width=\textwidth]{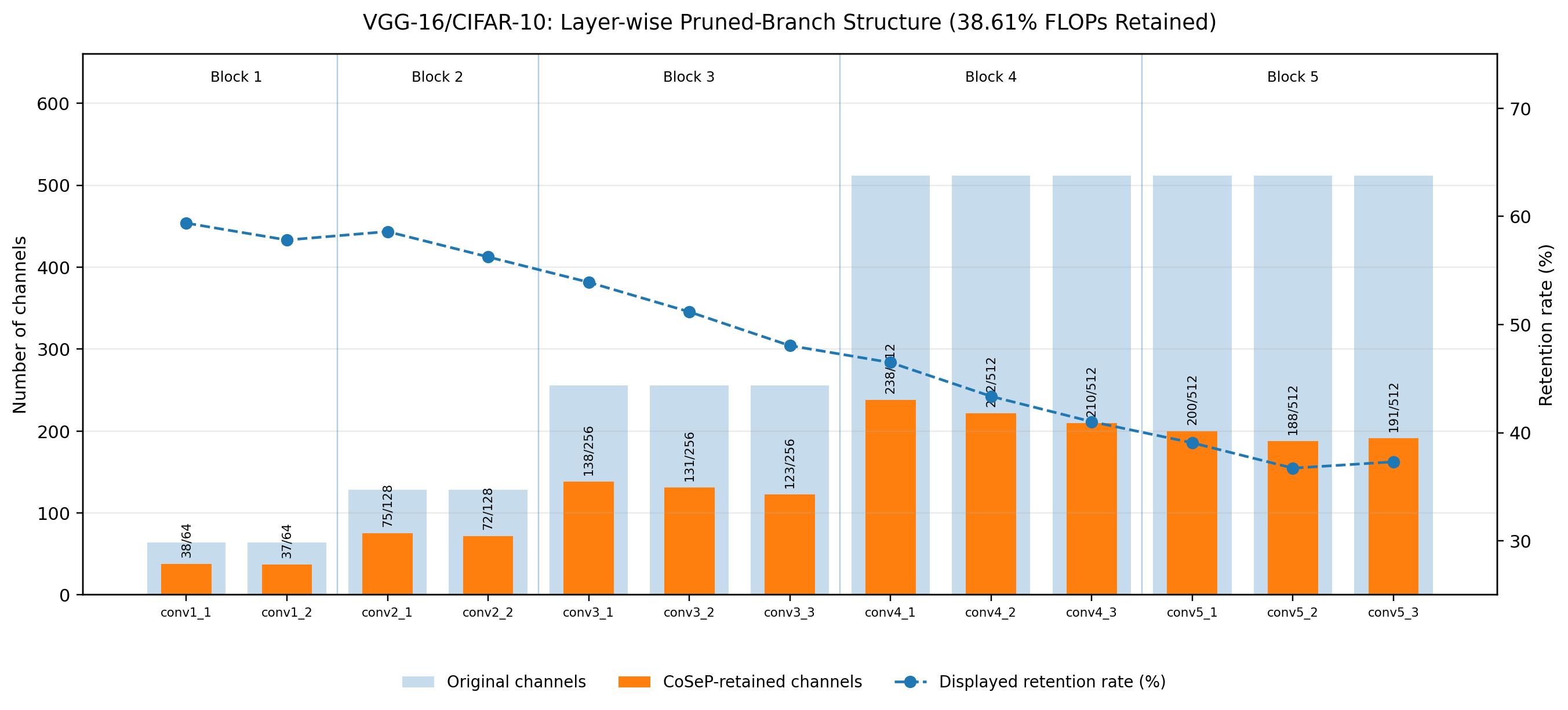}
    \caption{\textbf{Layer-wise channel retention for CoSeP on VGG-16/CIFAR-10 (38.61\% FLOPs retained).}
    Light-blue bars show original channels and orange bars show retained channels; labels report retained/original channel counts. The dashed curve is the retention rate, and vertical separators mark VGG blocks. Retention generally decreases with depth, with local non-monotonicity in the final block; all ratios are selected automatically by the MSS/Kneedle criterion.}
    \label{fig:vgg16-branch-structure}
\end{figure}

Figure~\ref{fig:vgg16-branch-structure} visualizes the architecture selected automatically for VGG-16. Its nonuniform profile shows that the MSS/Kneedle criterion adapts the retained cardinality by layer rather than imposing a single global schedule.

\begin{figure}[H]
    \centering

    \begin{minipage}[t]{0.49\textwidth}
        \centering
        \includegraphics[width=\textwidth]{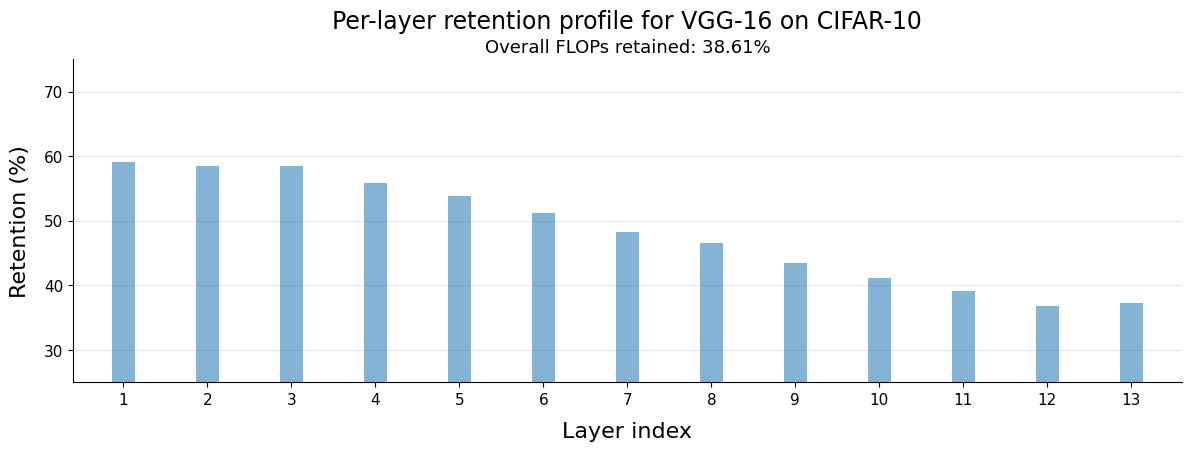}
    \end{minipage}
    \hfill
    \begin{minipage}[t]{0.49\textwidth}
        \centering
        \includegraphics[width=\textwidth]{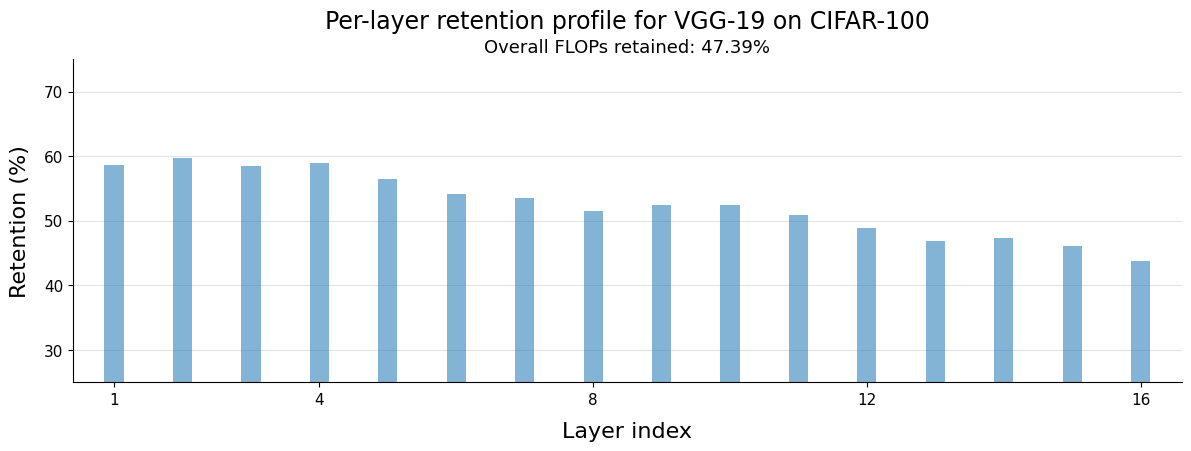}
    \end{minipage}

    \vspace{0.5em}

    \makebox[\textwidth][c]{%
        \begin{minipage}[t]{0.75\textwidth}
            \centering
            \includegraphics[width=\textwidth]{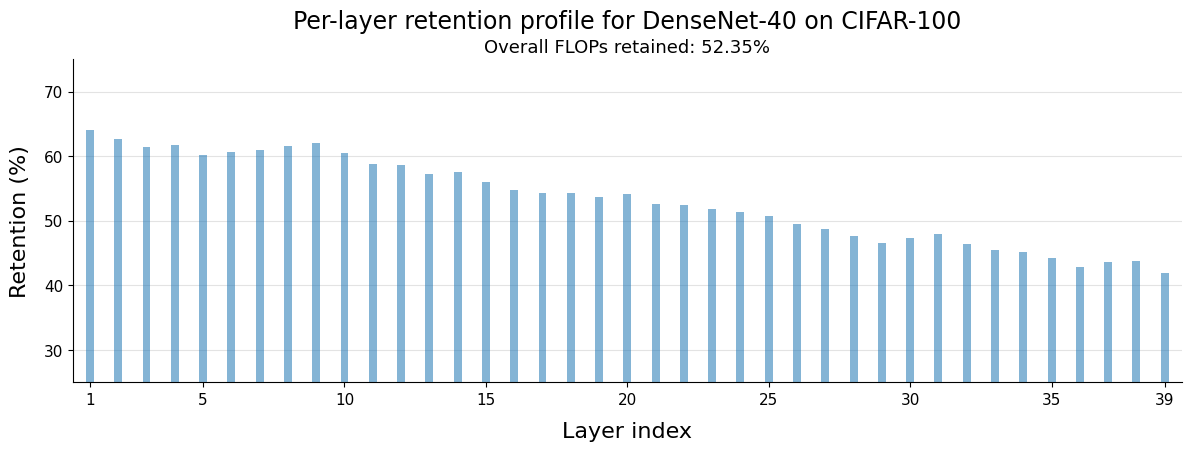}
        \end{minipage}
    }
    \caption{\textbf{Per-layer channel retention rate analysis.} Retention generally decreases with depth, although local fluctuations reflect architecture-specific redundancy patterns. The pattern emerges entirely from the MSS knee criterion without manual specification.}
    \label{fig:per-layer-retention-overview}
\end{figure}

Figure~\ref{fig:per-layer-retention-overview} compares the automatically selected profiles across VGG-16/CIFAR-10, VGG-19/CIFAR-100, and DenseNet-40/CIFAR-100. The stronger depth-dependent compression in the VGG models and the less regular DenseNet profile indicate that CoSeP responds to architecture-specific redundancy rather than transferring a fixed layer-wise pattern.

\end{document}